# Development of Bidirectional Series Elastic Actuator with Torsion Coil Spring and Implementation to the Legged Robot

Yuta Koda[1], Hiroshi Osawa[1], Norio Nagatsuka[1], Shinichi Kariya[1], Taeko Inagawa[1], Kensaku Ishizuka[1]
[1]Sony Interactive Entertainment

**Abstract— Many studies have been conducted on Series Elastic Actuators (SEA) for robot joints because they are effective in terms of flexibility, safety, and energy efficiency. The ability of SEA to robustly handle unexpected disturbances has raised expectations for practical applications in environments where robots interact with humans. On the other hand, the development and commercialization of small robots for indoor entertainment applications is also actively underway, and it is thought that by using SEA in these robots, dynamic movements such as jumping and running can be realized. In this work, we developed a small and lightweight SEA using coil springs as elastic elements. By devising a method for fixing the coil spring, it is possible to absorb shock and perform highly accurate force measurement in both rotational directions with a simple structure. In addition, to verify the effectiveness of the developed SEA, we created a small single-legged robot with SEA implemented in the three joints of the hip, knee, and ankle, and we conducted a drop test. By adjusting the initial posture and control gain of each joint, we confirmed that flexible landing and continuous hopping are possible with simple PD position control. The measurement results showed that SEA is effective in terms of shock absorption and energy reuse.**

**This work was performed for research purposes only.**

## I. Introduction

In recent years, there has been proactive development of robotic technology as a substitute for human labor, and there are high expectations for its application in tasks such as operations in extreme environments where direct human intervention is challenging, as well as the optimization of precision assembly operations [1][2]. These robots tend to be large and heavy, as they utilize high-torque and high-rigidity actuators to stably exert forces equal to or greater than those of humans. On the other hand, small and lightweight robots are being actively developed in the research and entertainment fields, and the possibility of providing new entertainment and companionship in our daily lives is expanding [3][4]. We have been researching and developing a small humanoid robot (EVAL-03) to conduct a feasibility study of such an entertainment robot [5][6]. Although EVAL-03 is small with a height of approximately 0.3m and a weight of 1.5kg, it has 26 joints, allowing it to perform expressive movements such as imitating human motion (Fig. 1). However, to perform agile movements such as running and jumping, the lack of shock absorption capability and insufficient actuator output torque are remaining issues. In addition, robots used for entertainment purposes, where there is a high possibility of interaction with users and the other robots, need flexibility to respond robustly to various external disturbances.

There have been three main approaches of actuator technology to make robots more agile and flexible.

Hydraulic actuators have extremely high energy density and excellent impact resistance. In addition, because output and speed can be controlled quickly and easily, it is used in state-of-the-art robots such as IHMC's Nadia [7], Boston Dynamics' Atlas [8] and Big Dog [9]. However, the hydraulic actuator becomes large because it requires complex mechanisms such as a hydraulic cylinder, a pump, and an oil tank. Additionally, there is a problem that the environment in which it can be used is restricted due to the risk of oil leakage.

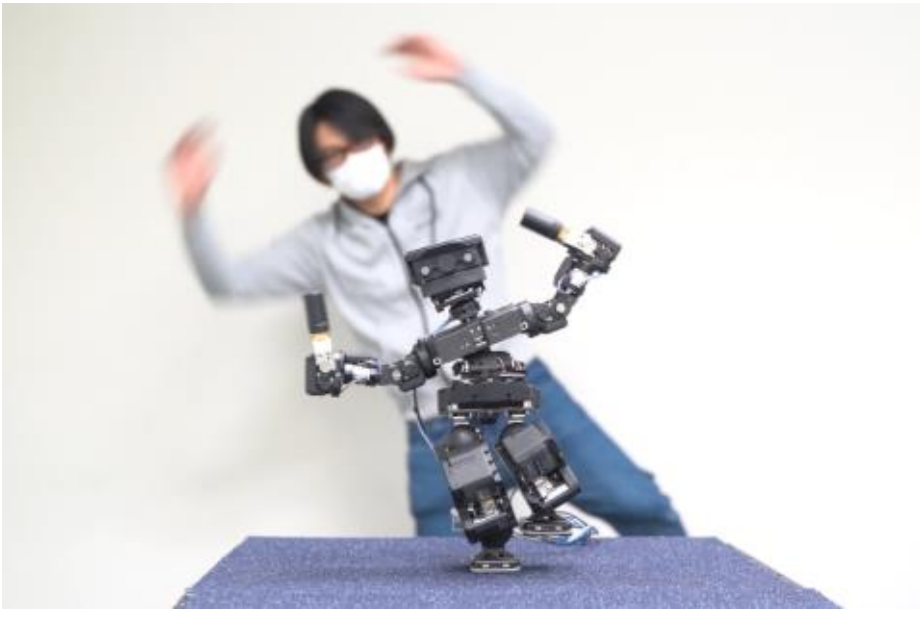

Figure 1. EVAL-03 imitating human motion

A direct drive actuator is a system in which an electromagnetic motor drives the actuator directly without using a decelerator. Since there is no reduction gear mechanism, it has high response and high efficiency, and because there is no backlash, precise positioning control is possible. Furthermore, since the noise level is low, it is an extremely suitable drive mechanism for entertainment robots. MIT's Cheetah [10], despite not being a fully direct-drive mechanism, achieves highly agile movements such as running and jumping using an actuator with an extremely low gear ratio. However, achieving high torque without a gearbox necessitates enlarging the motor, making practical application to small robots still challenging.

Series Elastic Actuator (SEA) consists of a motor (and decelerator), an elastic element, and a sensor that measures the amount of deformation of the elastic element. The force generated by SEA can be estimated from the amount of deformation of the elastic element and its spring constant, making it possible to conduct highly precise force control [11]. Due to its ability to add mechanical compliance, SEA is also superior in terms of shock absorption and storage of energy. Taking advantage of these benefits, research is underway on mechanical structures and control methods that consider the interactions between humans and robots [12][13], as well as improving the mobility of the robots themselves [14]. However, assuming that robots are about the same size as humans, making it difficult to apply the technology to small entertainment robots due to their structure. On the other hand, taking advantage of the fact that SEA is applicable to actuators with reduction gears, there are studies that achieve both downsizing and high agility of robots by using small geared motors [15][16]. UC Berkeley's Salto can jump up to 1m height using a SEA that combines a geared motor and a coil spring, despite its small size of only 0.1kg. Furthermore, the ability to perform continuous jumping utilizing the impact force during landing represents one of the examples that fully maximizes the characteristics of SEA. However, both Salto [15] and Jumper [16] are specialized for jumping motions, and SEA's springs only act in the compression direction, which poses issues in terms of versatility.

In this work, we assume future technological applications to EVAL-03 and proceed with the development and evaluation of a compact SEA with the following features:

- Capable of performing SEA functions in both rotational directions so that its movement as an entertainment robot will not be restricted.
- Modularize the actuator to enhance versatility.
- Using low-stiffness springs to match the mass of small and lightweight robots.
- Has the specifications necessary to operate a robot of the size equivalent to EVAL-03.

In this paper, we first describe the development and evaluation of the SEA unit itself. Subsequently, we proceeded with the development and evaluation of a single-legged robot using the developed SEA in each joint, verifying the effectiveness of SEA in a robotic system.

## II. Designing of SEA

### A. Type of SEA

The wide applicability of SEA has led to investigations into its potential usage in various fields, resulting in numerous structural variations customized for specific use cases. SEA can be classified into Force-sensing SEA (FSEA), Reaction Force-sensing SEA (RFSEA), and Transmitted Force-sensing SEA (TFSEA) depending on the position of the elastic element relative to the transmission [17].

FSEA is a SEA composed in the following order: motor, reduction gear, elastic elements, and displacement sensor. The force applied to the end effector of the reduction gear can be measured as the displacement of the spring. Since the force generated by FSEA is equivalent to the spring force of an elastic element, force control can be easily performed. In addition, when external forces are applied, the elastic elements can protect the transmission and motor by filtering the impact.

RFSEA has a structure in which a spring is placed between the motor and transmission or between the motor and the ground. Although it has the advantage of higher torque transmission efficiency than FSEA, it requires a structure that allows the motor or transmission to move as the spring deforms, making the actuator more complex and larger than FSEA.

TFSEA is configured to measure the transmission torque of the gear train by placing a spring inside the transmission. Compared to FSEA, torque transmission efficiency is higher, but the load applied on the gear ahead of the spring is higher in response to external impact.

In this research, we designed a new actuator based on FSEA because we prioritized impact resistance and miniaturization rather than the transmission efficiency of the actuator.

### B. Mechanical design

The specifications of the SEA are shown in Table 1, and the configuration diagram is shown in Fig. 2. As mentioned earlier, we assumed that the SEA technology examined will be introduced into EVAL-03 in the future. It is desirable that the driving specifications of SEA are equivalent or greater, and highly compatible with systems used in EVAL-03. Therefore, we have decided to reuse the motor, reduction gear, and drive circuit used in EVAL-03. The EVAL-03 actuator's output shaft (Axis-A) and the SEA's rotating shaft (Axis-B) are connected by a timing belt, and a coil spring is installed on the rotating shaft as an elastic element. TMR sensors A and B and magnets are installed on the rotation axes of Axis-A and Axis-B as angle sensors. It is possible to measure the deformation of the coil spring, i.e., the generated torque, from the difference in the rotational angles measured by TMR sensors A and B.

Fig. 3 shows the component configuration of the coil spring and SEA output shaft. A coil spring was placed between the timing belt pulley and the output shaft. Generally, coil springs are fixed by applying a preload to the spring. However, with this method, spring force can only be generated in the direction in which the spring flexes. By using a method of fixing the legs of the coil spring to each component with set screws, we created a structure where spring force acts in both rotational directions. Furthermore, in the manufacturing process of coil springs, the adoption of an open-wound configuration, rather than a close-wound, is considered to reduce hysteresis caused by friction during the deformation of the spring. The movable range of the coil spring is provisionally set to ±40 degrees, and stoppers are installed to prevent permanent deformation of the spring. The spring constant of the coil spring was set to 0.015 N·m/deg, that is 0.6 N·m at the maximum deflection angle of ±40 degrees. This value is approximately 50% of the actuator's stall torque of 1.33 N·m and is within the range where the actuator will not be damaged even if the maximum torque of the spring is continuously applied.

TABLE I. SPECIFICATION OF SEA

| Parameter | Units | Data |
|---|---|---|
| Input Voltage | V | 15 |
| Stall Torque | N·m | 1.33 |
| Torque Constant | N·m/A | 0.48 |
| No Load Speed | rpm | 156.5 |
| Gear Ratio | - | 1:61.4 |
| Backlash | deg | 0.22 |
| Back Drive Torque | N·m | 0.135 |
| Mass | g | 117.8 |

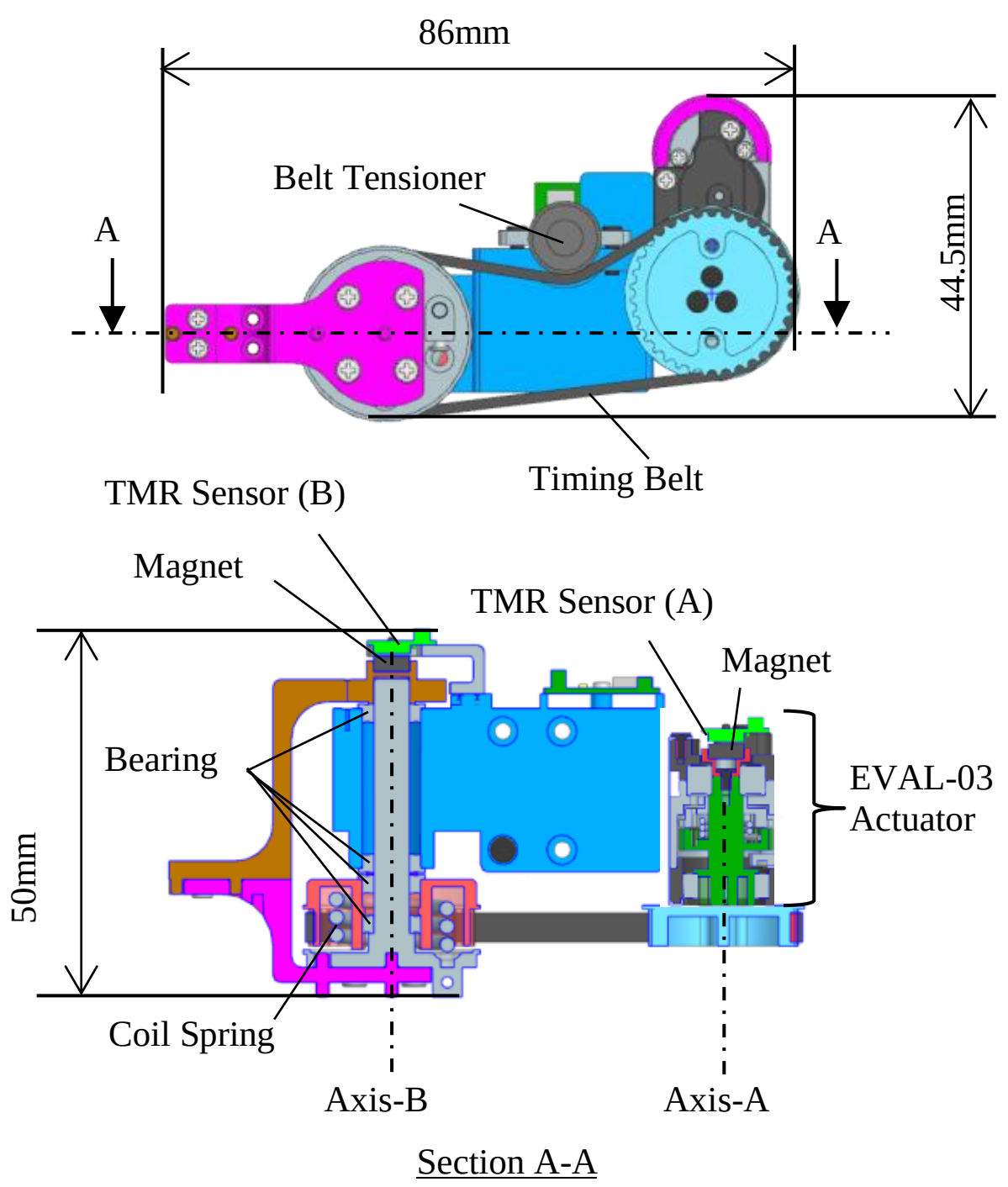

Figure 2. Configuration diagram of SEA.

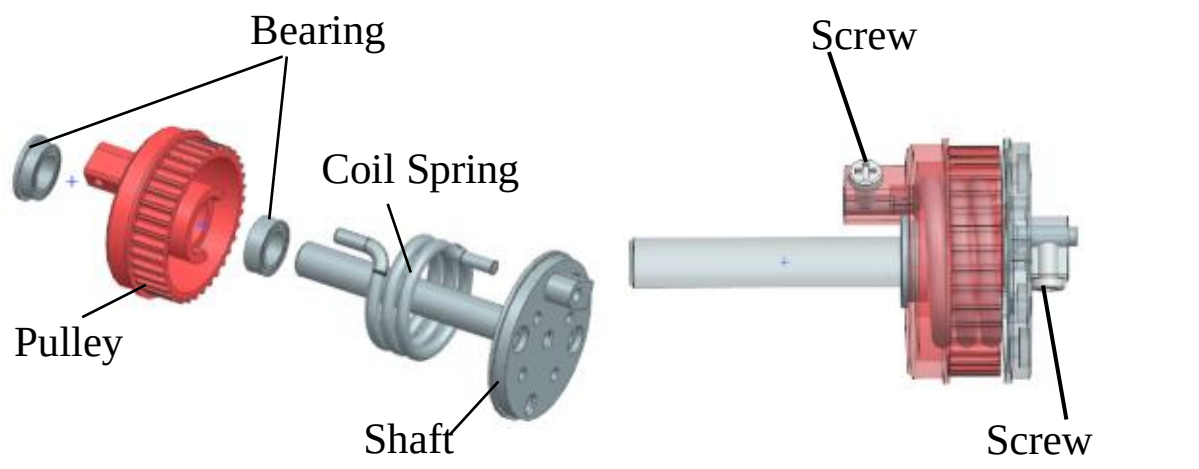

Figure 3. Component configuration of SEA output shaft and coil spring.

## C. *Spring constant measurement*

For measuring the spring constant of the developed SEA, we constructed a measurement fixture as shown in Fig. 4. The SEA with a pulley attached to the output shaft was fixed with a clamp, and the rotational angle of the output shaft was measured when a weight load was applied in both rotational directions. During measurements, a lock gear is installed instead of the motor to prevent the input shaft of the actuator from rotating by the weight load. To verify the spring characteristic across the entire movable range of the SEA, incremental loading and unloading were conducted in each rotational direction.

The measurement result is shown in Fig. 5. The relationship between torque and deformation of the spring shows relatively small hysteresis and linear behavior in both rotational directions, indicating the capability for high-precision force measurement. The measured spring constant is approximately 0.012 N·m/deg, nearly identical to the design value, confirming that the spring is functioning as designed.

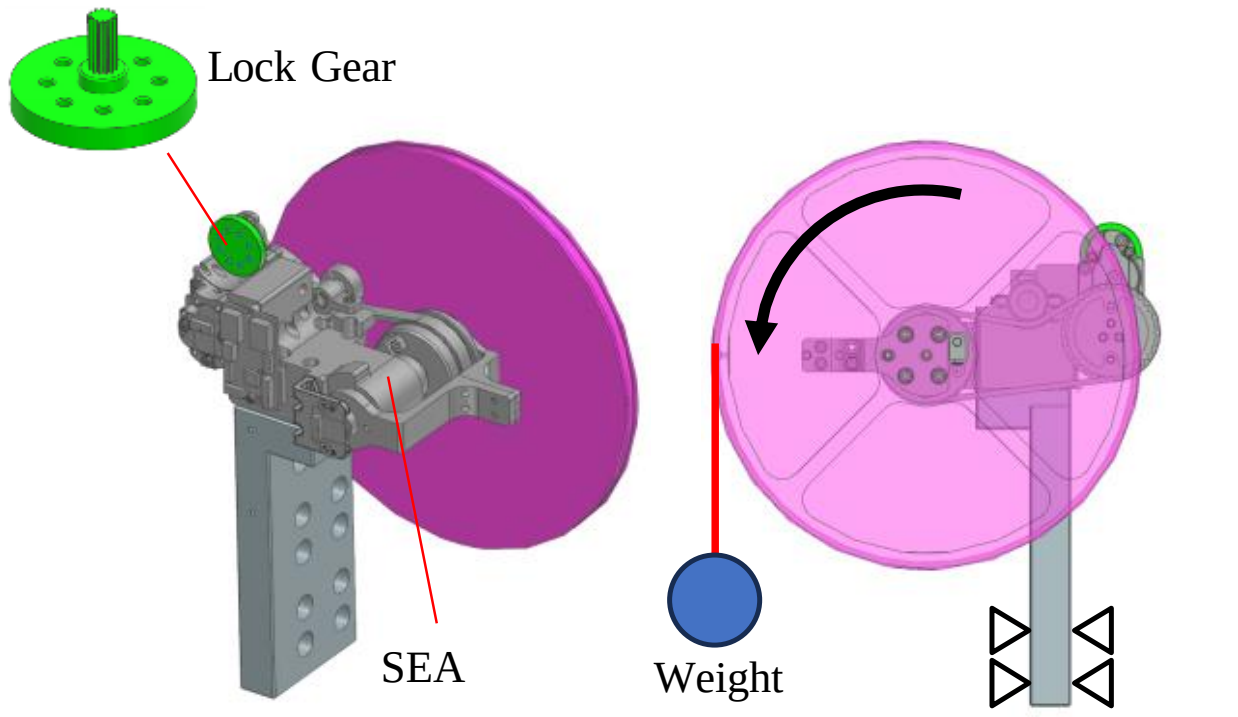

Figure 4. Fixture for measuring the spring constant of SEA. Deformation of the spring was measured as the rotational angle of the wheel by the TMR sensor.

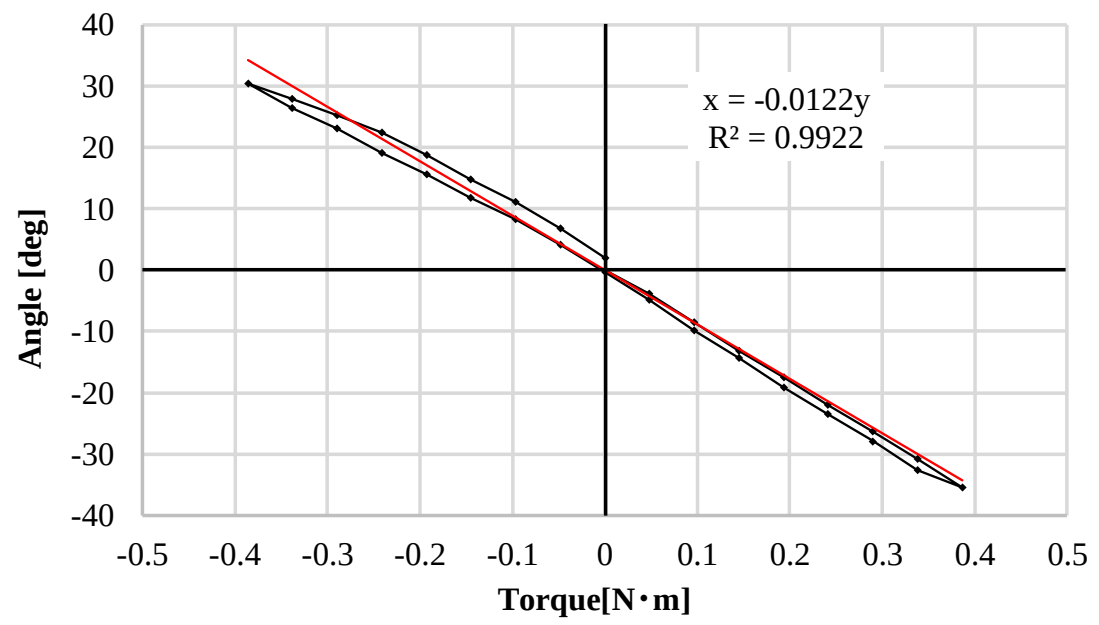

Figure 5. Measurement result of the spring constant of SEA.

# III. IMPLEMENTATION TO LEGGED ROBOT

## A. *Mechanical Structure*

To verify the effectiveness of the developed SEA, we created a single-legged robot implementing SEA. To increase compatibility with EVAL-03, the length of each link of the leg, the movable range of each joint, and the overall mass are designed based on the values of EVAL-03. Fig. 6 shows the mechanical structure of the single-legged robot, and Table 2 shows the movable range and mass of each joint. The created single-legged robot has three degrees of freedom: Hip, Knee, and Ankle, and SEA was implemented at each joint. To achieve miniaturization of the legged robot, we have replaced components corresponding to the frame and arm parts of the SEA unit shown in Fig. 2 with the structure parts of the legged robot. Regarding the sole of the robot, we used a rigid plate in this work, but it is also possible to replace it with a sole equipped with a center of gravity sensor that is similar to EVAL-03, or a sole with built-in shock absorption.

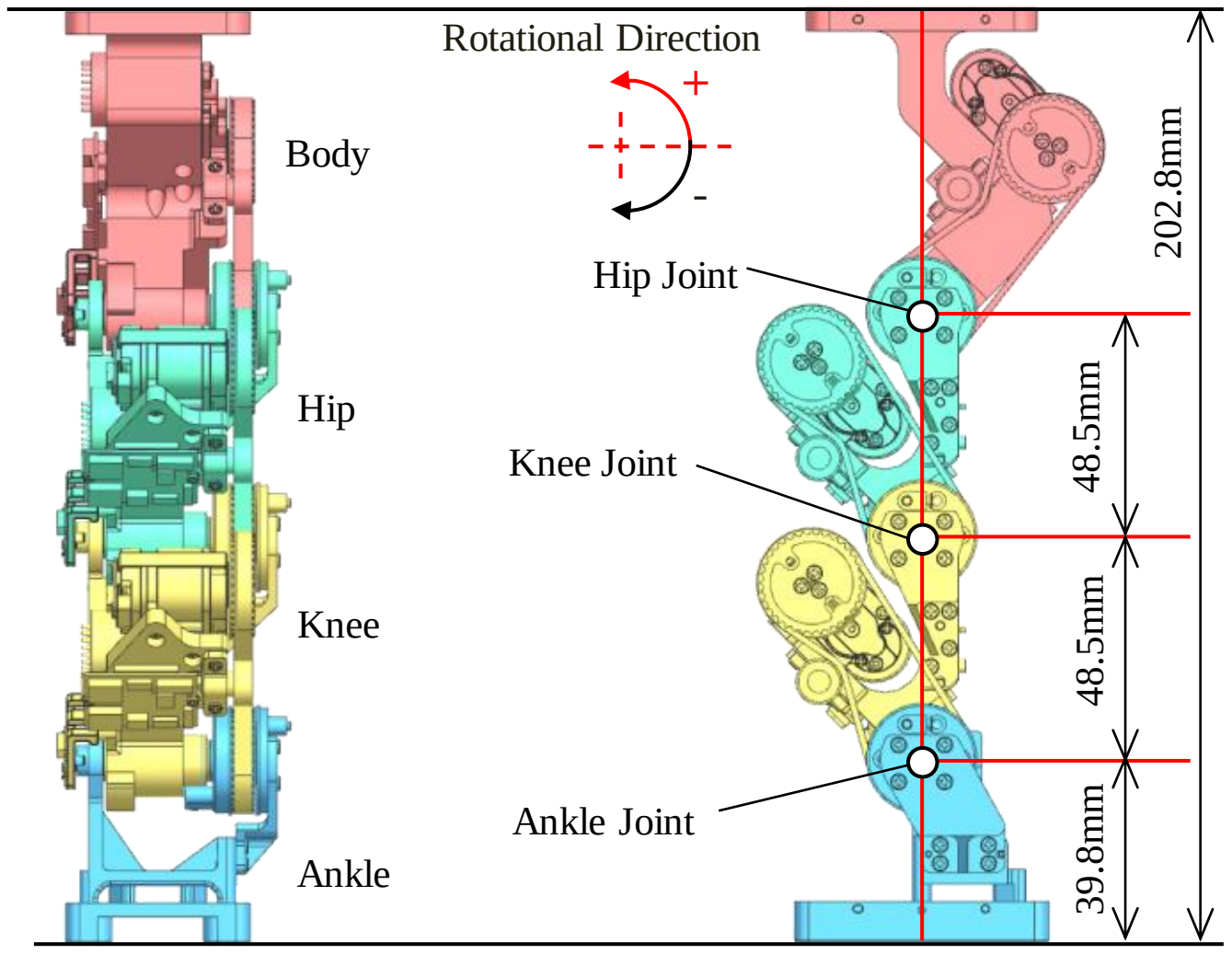

Figure 6. Mechanical structure of the developed single-legged robot. The actuators that drive each joint are installed in the section immediately above that joint. e. g. , the actuator responsible for driving the Hip joint is installed in the Body section.

TABLE II. MOVABLE RANGE AND MASS OF EACH JOINT

| Link | Movable Angle [deg] | Mass [g] |
|---|---|---|
| Body | - | 132.0 |
| Hip | -93~37 | 107.4 |
| Knee | -1.5~127 | 107.0 |
| Ankle | -82~37 | 67.6 |

### B. *Measurement of natural frequency*

The natural frequency of each joint was measured to obtain information to determine the control parameters of the single-legged robot. To simplify the measurement, we replaced the SEA springs with rigid parts so that joints other than the ones being measured could be treated as rigid bodies without springs. The motors were fixed with the lock gears used in the spring constant measurement experiment described previously. During the measurement, the Body part was secured to a stand to prevent the sole from contacting the ground, and the targeted joint was manually flicked to induce free vibration.

The measurement results are shown in Fig. 7. From Fig. 7a, it can be observed that the amplitude and settling time increase in the order of Hip, Knee, and Ankle, which have larger inertia. Additionally, the Fourier transform results in Fig. 7b shows that the natural frequency of Hip is the lowest, increasing in the order of Knee and Ankle. These results are related to the adjustment range of control gains at each joint and were taken into consideration to avoid unexpected oscillation.

## IV. EXPERIMENTS

### A. *Experimental Setup*

In this research, we will conduct verification experiments focusing mainly on the shock absorption performance of SEA and the efficiency of energy use utilizing the force stored in springs. To understand the characteristics of the SEA, we performed a drop test with the posture fixed using simple PD position control and observed the behavior when the elastic elements of the SEA were passively deformed by the impact of landing. Fig. 8 shows the drop test fixture. The single-legged robot is fixed to a linear slider by its Body part, which restricts movement other than in the vertical direction.

The PD control gain of each joint was set with reference to the measurement results of the natural frequencies in Fig. 7(b). However, the Hip joint is more prone to oscillations compared to the Knee and Ankle joints, making it difficult to stabilize the motion only with gain adjustments. This is attributed to the lower natural frequency of the Hip joint, and it can be mitigated by increasing the spring constant and adjusting the weight balance of each joint. In this study, for the Hip joint, we replaced the SEA's coil spring with a rigid component and evaluated the overall performance. Optimization of the spring constant of each joint based on the natural frequency and inertia is a future topic.

The falling height was fixed at 70mm, and drop tests were conducted by varying the PD control gains of the three joints and the initial angle of each joint. In this paper, we will discuss two patterns of test results: one involving a landing motion that absorbs the impact of the fall and another involving a continuous hopping motion that utilizes the recoil from the falling impact.

### B. *Landing motion*

While it is recommended to view the results via supplementary video, snapshots of landing motion are shown in Fig.9. The significant flexion of the knee joint helps absorbing the impact, allowing for a smooth and stable landing.

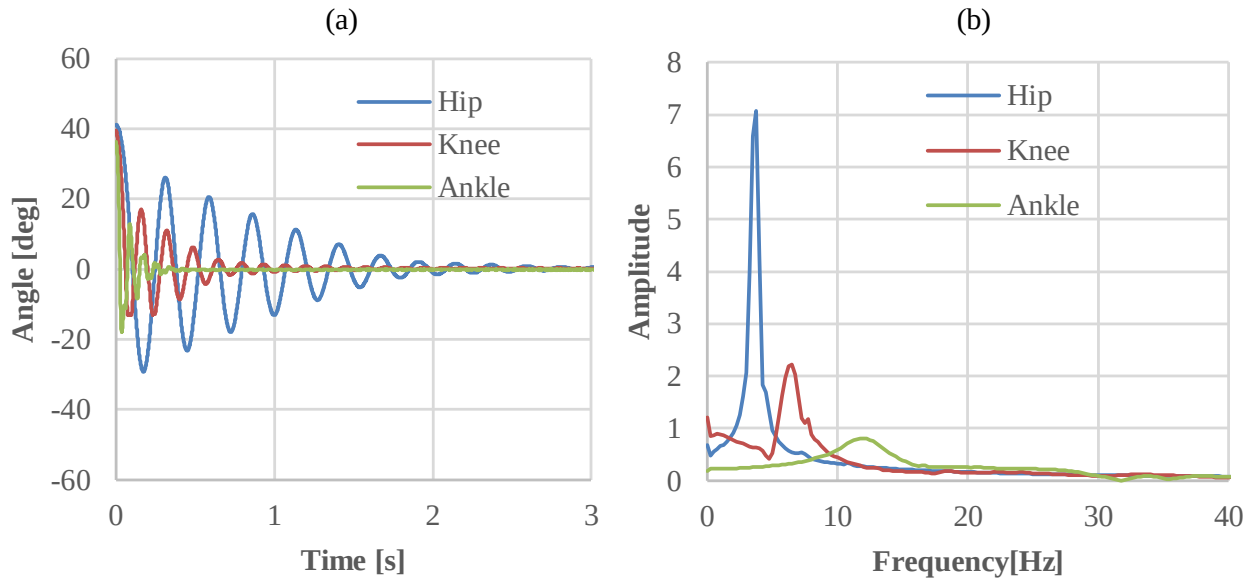

Figure 7. Measurement result of the natural frequency of each joint in the developed single-legged robot.

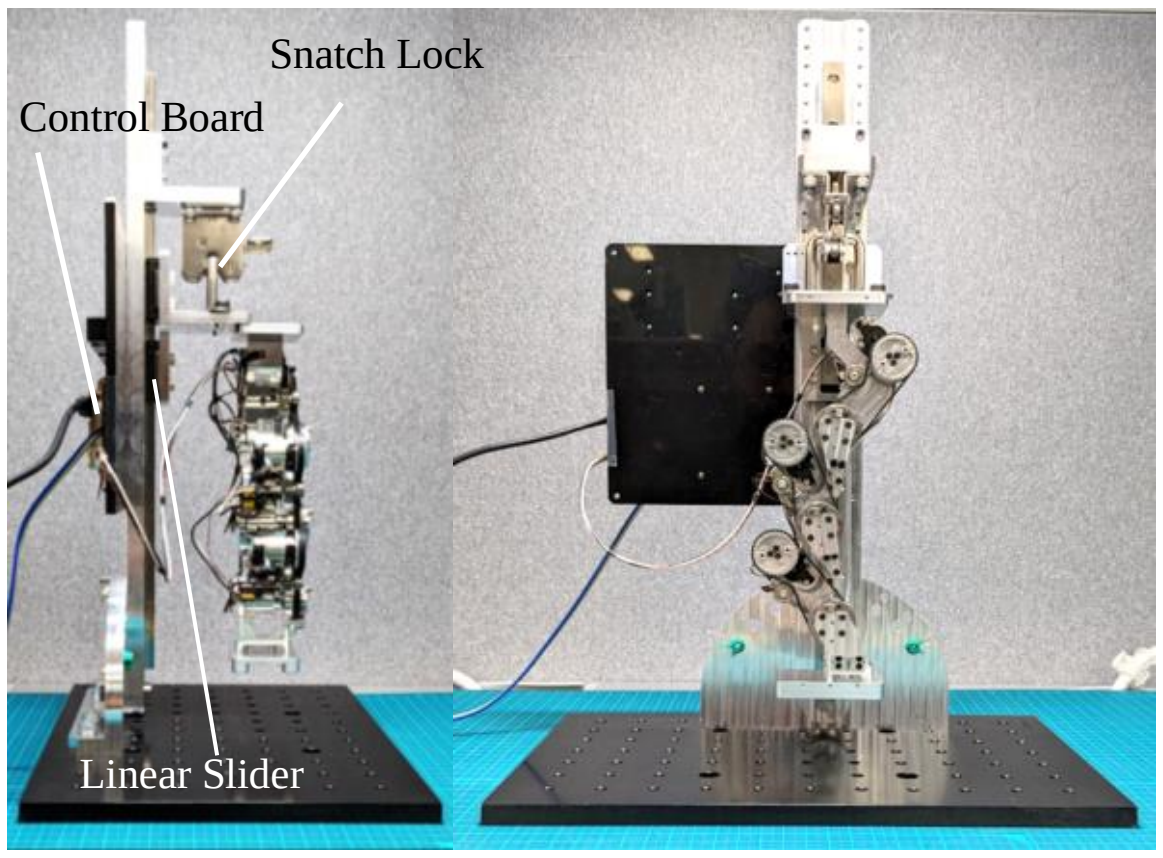

Figure 8. Drop test fixture with the developed single-legged robot. To avoid affecting the robot's operation, the cables from the robot to the control board are securely fastened to the fixture pillar with sufficient slack.

For comparative verification, the same test was conducted with all joints' SEA springs fixed. The measurement results of the Ankle and Knee joints during the landing motion are shown in Fig. 10 and 11. In Fig. 10(a) and 11(a), when all the SEA springs are fixed, the rotational angles of the Ankle and Knee joint and corresponding SEAs angles are nearly identical, and the peak current of the actuators due to the landing impact are approximately -400mA for Ankle joint and -850mA for Knee joint. In the condition where the SEAs springs are active, Fig. 11(b) shows that the Knee joint deforms by approximately 20 degrees at landing to absorb the impact. Furthermore, with the peak current of the actuator driving Knee joint decreasing to around -600mA, it can be inferred that the load on the actuator has decreased by approximately 30% compared to when the spring was fixed. From these results, it is confirmed that the SEA spring is effective in shock absorption.

Comparing Figures 10(b) and 11(b), the change in angle due to the impact during landing is smaller for the Ankle joint compared to the Knee joint. This is because intentionally adjusting an initial posture to lands from the heel has led to a reduction of the load to the Ankle joint. Conversely, adjusting an initial posture to lands from the toe leads to a significant flexion of the Ankle Joint, enabling the utilization of falling energy for jumping. Further details will be explained in the hopping motion section later.

When comparing the time from landing to stabilization, it takes about 0.1 seconds when the spring is fixed, whereas in the case of a free spring, the Knee joint takes about 0.2 seconds due to the influence of minor bouncing caused by spring recoil.

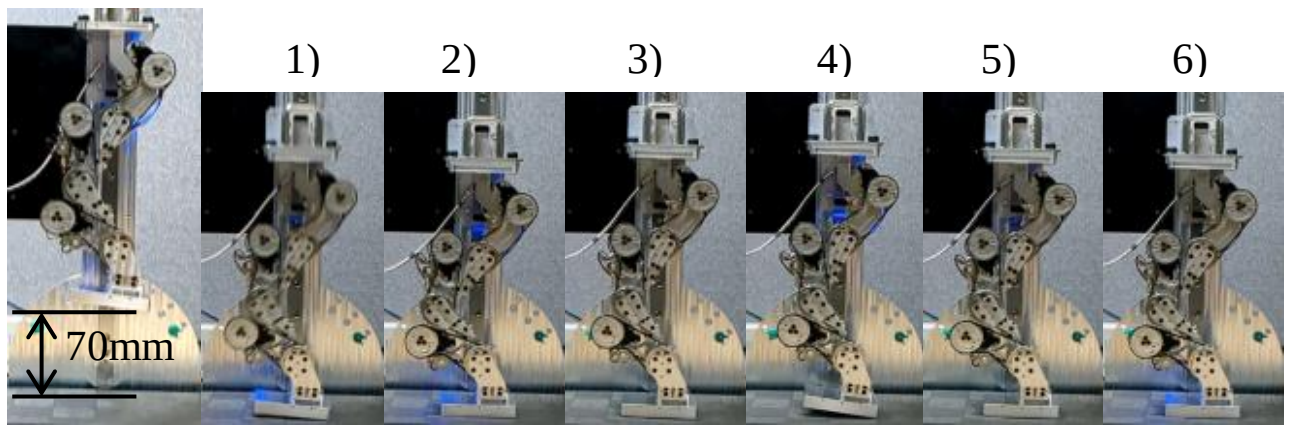

Figure 9. Snapshots of landing motion.

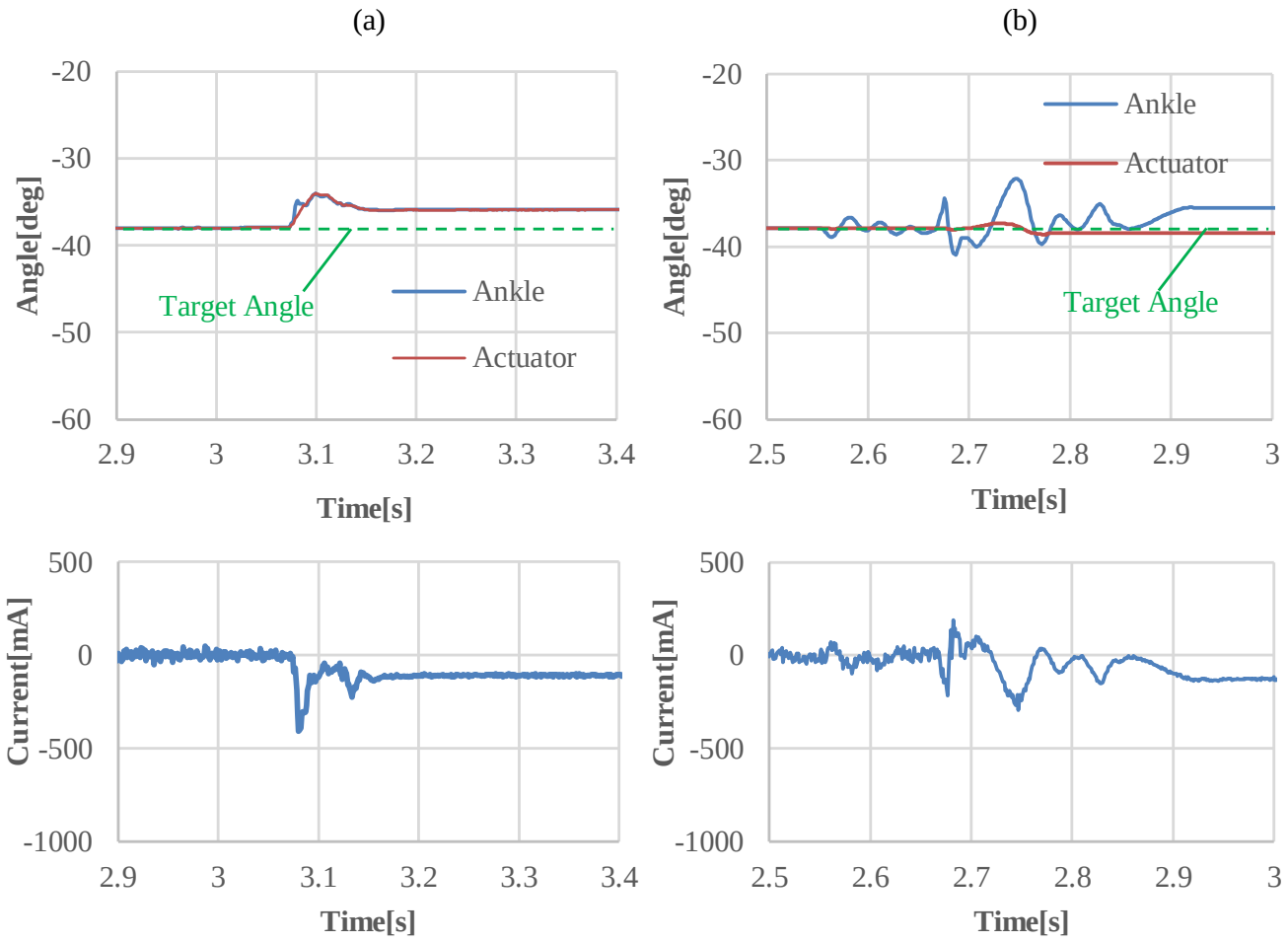

Figure 10. Measurement result of Ankle joint at the landing motion. Figure 10 (a) is the result with the springs of each joint's SEA fixed, while Figure 10 (b) illustrates the result with the springs of the Ankle and Knee SEAs activated.

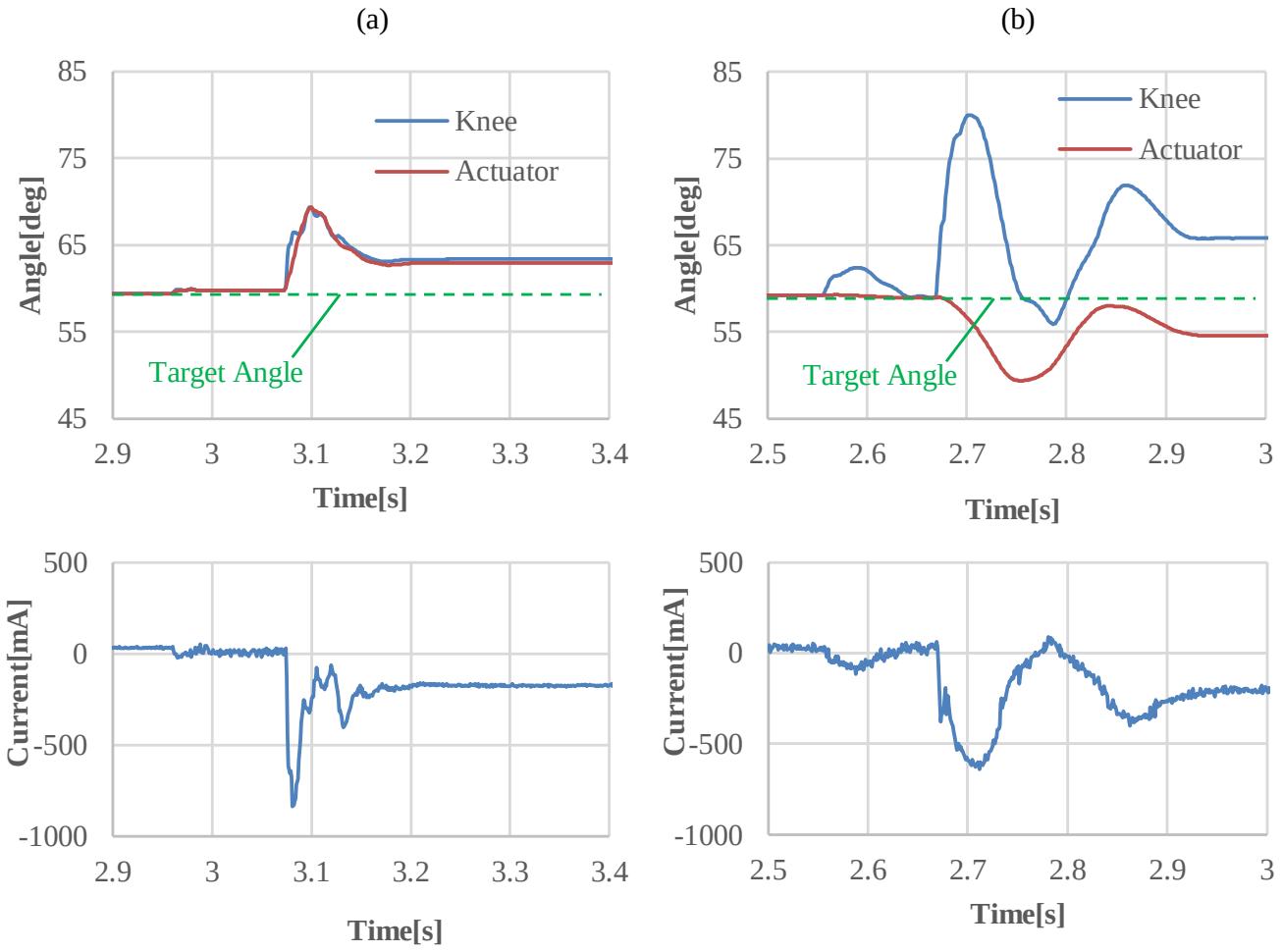

Figure 11. Measurement result of Knee joint at the landing motion. Figure 11 (a) is the result with the springs of each joint's SEA fixed, while Figure 11 (b) illustrates the result with the springs of the Ankle and Knee SEAs activated.

The integrated current flow of the Knee joint from the landing to complete stabilization is approximately 30A/s when the spring is fixed, while it increases to around 70A/s in the state with a free spring. From these results, it is evident that the introduction of SEA worsens performance in terms of stabilization time and power consumption. However, we believe that the performance can be improved by introducing damping control utilizing sensor information on joint angles and actuator angles dynamically. We will continue further investigation and consideration as a future issue.

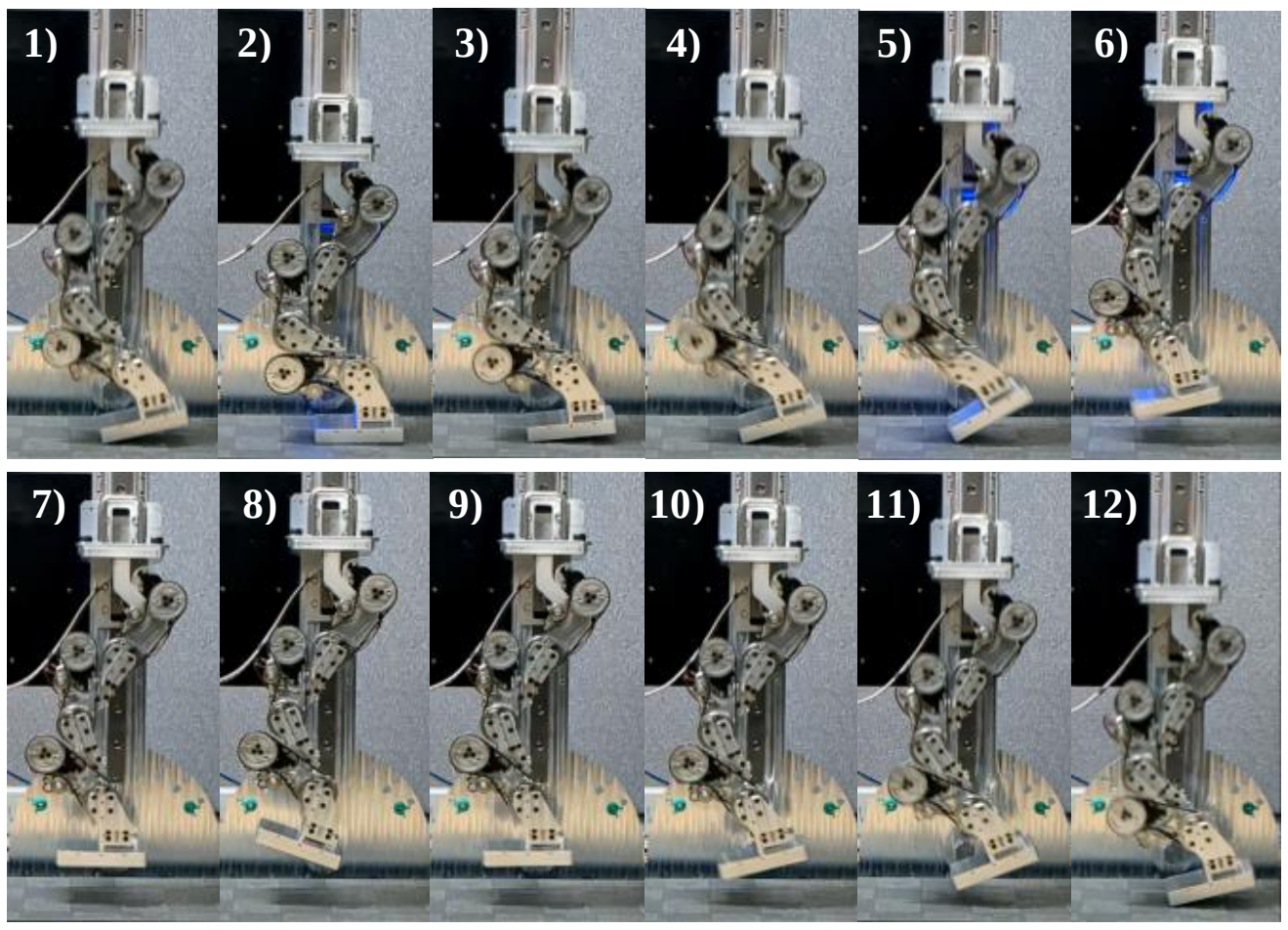

Figure 12. Snapshots of hopping motion.

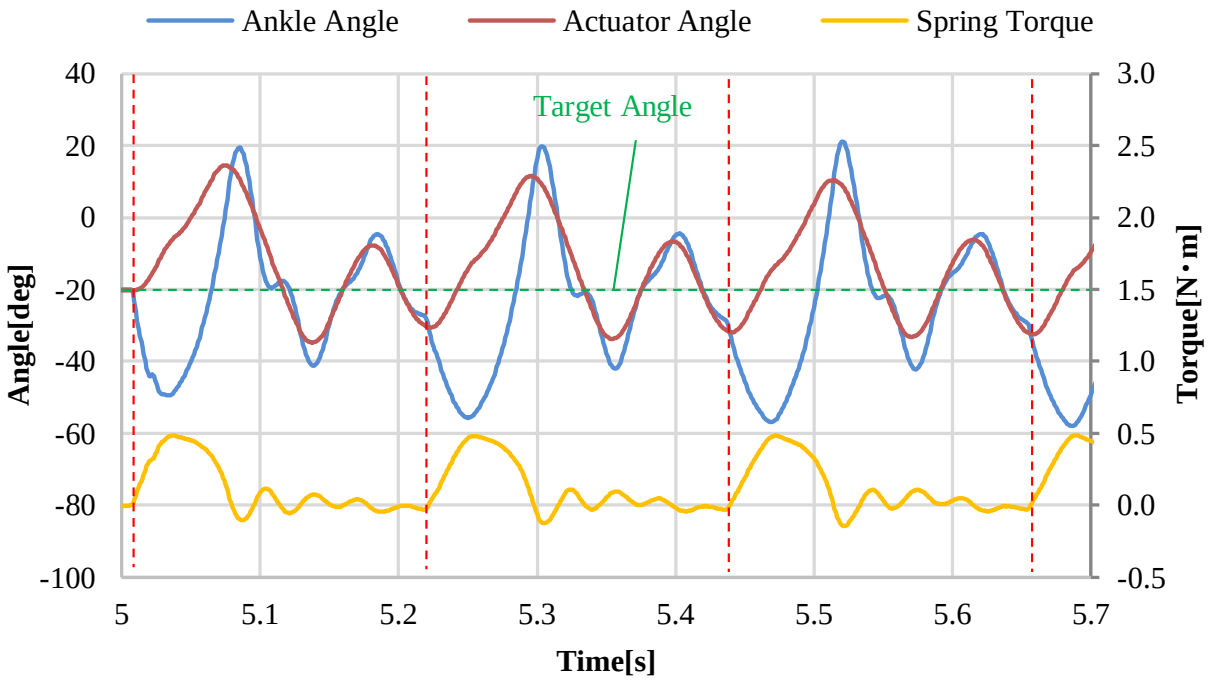

Figure 13. Measurement result of Ankle joint during hopping motion.

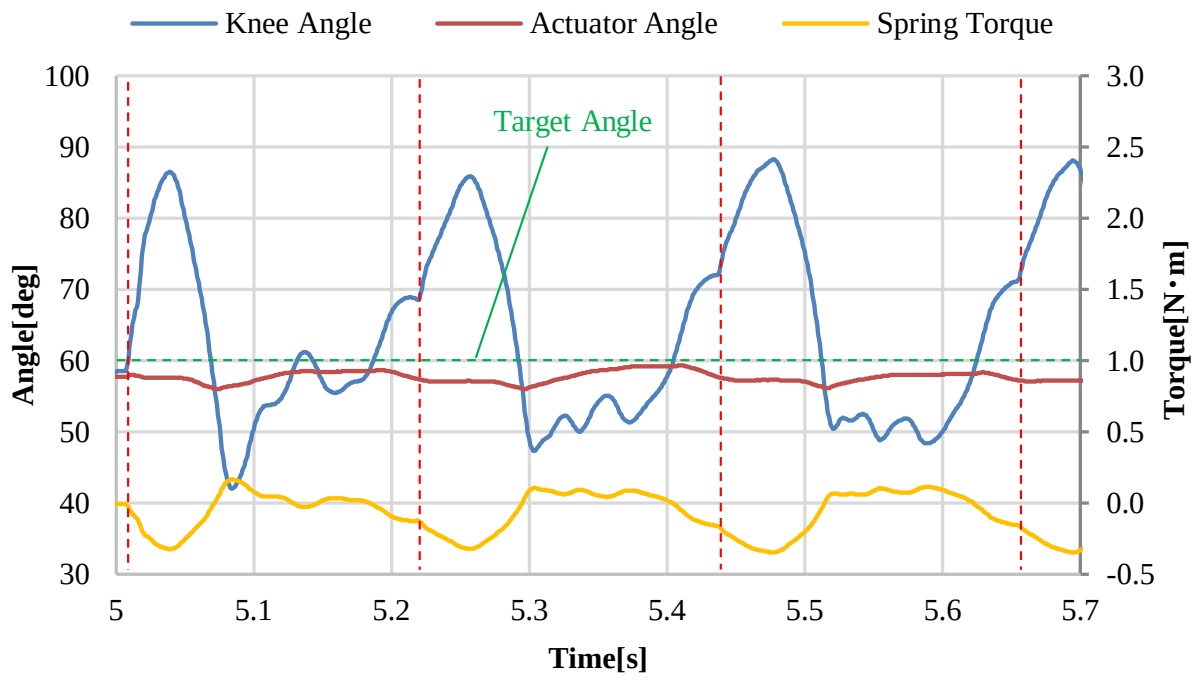

Figure 14. Measurement result of Knee joint during hopping motion.

## C. *Hopping motion*

Fig. 12 shows snapshots of the hopping motion, while Fig. 13 and 14 illustrates the measurement results of three consecutive hopping motions. The red dashed line in Fig. 13 and 14 indicates the time when the toe of the legged robot touched down. We recommend watching the supplementary video for hopping motion as well as the results for landing motion.

From Fig. 12, it is evident that the posture of the legged robot is nearly the same when it first lands (Fig. 12-1) and when it lands again (Fig. 12-12). The joint angle at the time when the toe of the legged robot touches down in Fig. 13 and 14 are almost equal, indicating a consistent posture during landing. This is due to the nearly identical motion cycles of each joint of the legged robot and the hopping motion, demonstrating stable and continuous hopping motion.

From Fig. 13 and 14, it can be observed that during hopping motion, the spring torques of the Knee joint and Ankle joint undergo significant changes simultaneously with the landing. This indicates that the legged robot performs hopping movements through the coordinated action of these joints. Particularly in the case of the Ankle joint, not only the spring torque but also the actuator undergoes substantial movement, enabling the generation of the necessary torque for hopping. This is possible due to the higher proportional gain set for the Ankle joint, which has the highest natural frequency among the joints. For the Knee joint, with a lower proportional gain setting, the actuator movement is minimal, but the spring torque from SEA contributes to the hopping motion. From the above results, we were able to demonstrate the effectiveness of SEA in hopping motion of a legged robot under the limited condition of simple PD position control.

## V. Limitation and Future work

As mentioned earlier, the technology of the created legged robot is expected to be transferred to EVAL-03, so the actuators and body size are designed to match EVAL-03 to facilitate future research. Therefore, the mechanical structure has not been optimized for miniaturization. We believe that by reviewing the reducer and power transmission structure, we can expect to further downsize the SEA unit and build a more versatile system. Additionally, since SEA with the same spring constant is used for all joints of the legged robot, it was difficult to stabilize the motion by simply adjusting the PD control gain. In the future, we will consider adding damper elements and optimizing the SEA spring constant to match the natural frequency and inertia of each part of the robot. Other considerations include creating a walking bi-pedal robot by connecting two single-legged robots, investigating the impact of changing sole size and rigidity on mobility, and extending SEA to joints other than the legs.

In controlling legged robots, we have focused on verifying the basic characteristics of the developed SEA, so we have only tried the drop test during simple PD position control. In the future, we will conduct the following studies to achieve more flexible and highly agile operations: characteristic analysis of SEA in walking and running motions, adaptation to disturbances using a combination of SEA's mechanical and actuator impedance control, force amplification, and reduction control synchronized with the deformation of the SEA spring.

## VI. Conclusions

In this work, a compact SEA unit utilizing a coil spring was developed, and its fundamental characteristics were measured. Additionally, a single-legged robot using the created SEA was developed, and drop tests were conducted. It was confirmed that landing motions and continuous hopping actions could be performed by absorbing the impact of the fall through the SEA. The created SEA exhibited a high linearity of spring characteristics in both rotational directions by ingeniously fixing the coil spring, enabling precise force measurements. In landing motion experiments of the legged robot, it was observed that the peak current of the motor decreased by approximately 30% when the spring of the SEA was active compared to when it was fixed, indicating that the SEA effectively reduced the load on the motor and the body by absorbing the impact of the fall. Furthermore, in hopping motion experiments, it was demonstrated that the Knee and Ankle joints with the implemented SEA worked in coordination to efficiently absorb and reuse falling energy, enabling continuous hopping.

## Acknowledgment

We would like to express our gratitude to our internal reviewers for the helpful feedback, together with Hiroshi Osawa and his team for their development and technical advice regarding EVAL-03.